\documentclass[10pt,journal,compsoc]{IEEEtran}

\ifCLASSOPTIONcompsoc
  \usepackage[nocompress]{cite}
\else
  \usepackage{cite}
\fi

\ifCLASSINFOpdf
\else
\fi
\usepackage{times}
\usepackage{latexsym}
\usepackage{amsmath}
\usepackage{amssymb}
\usepackage{amsmath}
\usepackage{amssymb}
\usepackage{enumitem}
\usepackage{graphicx}
\usepackage{multirow}
\usepackage{graphics}
\usepackage{amsthm,amsmath,amssymb}
\usepackage{makecell}
\usepackage[utf8]{inputenc} 
\usepackage{graphicx}
\usepackage{subcaption}
\usepackage{multirow}
\usepackage{amsmath}
\usepackage[ruled]{algorithm2e} 
\usepackage[dvipsnames]{xcolor}
\DeclareMathOperator*{\argmax}{arg\,max}

\begin{document}
%
\title{Relation Extraction from Biomedical and Clinical Text: Unified Multitask Learning Framework}
%
%
%
%

\author{Shweta Yadav,
        Srivatsa Ramesh,
        Sriparna Saha,
        and Asif Ekbal
\IEEEcompsocitemizethanks{\IEEEcompsocthanksitem All the authors are with the Department
of Computer Science and Engineering, Indian Institute of Technology Patna, Bihar,
India, 801103.\protect\\
E-mail: {(shweta.pcs14,sriparna,asif)}@iitp.ac.in
 }}

%
%

\markboth{Accepted for publication at IEEE/ACM Transaction on Computational Biology and Bioinformatics}%
{Shell \MakeLowercase{\textit{et al.}}: Bare Advanced Demo of IEEEtran.cls for IEEE Computer Society Journals}
%



\IEEEtitleabstractindextext{%
\begin{abstract}
\textbf{Motivation:} To minimize the accelerating amount of time invested on the biomedical literature search, numerous approaches for automated knowledge extraction have been proposed. Relation extraction is one such task where semantic relations between the entities are identified from the free text. In the biomedical domain, extraction of 
regulatory pathways, metabolic processes, adverse drug reaction or disease models necessitates knowledge from the individual relations, for example, physical or regulatory interactions between genes, proteins, drugs, chemical, disease or phenotype. \\
\textbf{Results:} In this paper, we study the relation extraction task from three major biomedical and clinical tasks, namely drug-drug interaction, protein-protein interaction, and medical concept relation extraction. Towards this, we model the relation extraction problem in a multi-task learning (MTL) framework, and introduce for the first time the concept of structured self-attentive network complemented with the adversarial learning approach for the prediction of relationships from the biomedical and clinical text. 
The fundamental notion of MTL is to simultaneously learn multiple problems together by utilizing the concepts of the shared representation.
Additionally, we also generate the highly efficient single task model which exploits the shortest dependency path embedding learned over the attentive gated recurrent unit to compare our proposed MTL models.
The framework we propose significantly improves over all the baselines (deep learning techniques) and single-task models for predicting the relationships, without compromising on the performance of all the tasks.  
\end{abstract}

}

\maketitle

\IEEEdisplaynontitleabstractindextext

%
\IEEEpeerreviewmaketitle

\ifCLASSOPTIONcompsoc
\IEEEraisesectionheading{\section{Introduction}\label{sec:introduction}}
\else
\section{Introduction}
\label{sec:introduction}
\fi
Owing to the rapid growth of the scientific literature, majority of the available biological facts remain concealed in the form of scientific literature. Over the last two decades, MEDLINE size has risen at a compounded annual growth pace of ~4.2 percent. 
MEDLINE currently holds more than $26,000,000$ records from $5639$ publications, which is more than $5$ millions than those indexed in $2014$ alone. This similar trend was also observed in case of the healthcare data. IBM reported that nearly $2.5$ quintillion bytes of the healthcare data are generated globally. Encapsulated within this unstructured text is an enormous amount of significant healthcare and the biomedical data, which are valuable sources of information for the Biomedical Natural Language Processing (BioNLP) domain.

As a consequence of the exponential rise \cite{lu2011pubmed,khare2014accessing} and complexity of the biological and clinical information, it is imperative to advance the automatic extraction techniques to assist biologist in detecting, curating and maintaining databases and providing automated decision support systems for the health professional. This has led to a rise in the interest of the BioNLP community to automatically detect and extract information from the scientific literature and clinical records \cite{yadav2016deep,yadav2018feature,yadav2017entity}. \\
\indent Relation extraction is one such task 
that aims to detect and characterize the semantic relationship between the biological/clinical entities. The relation types could vary depending upon the genres and domains, such as interactions between genes, proteins, drugs, or medical concepts (problems, treatments, or tests).\\
In this paper, we study the relation extraction (RE) on the most popular biomedical and clinical tasks, namely drug-drug interaction (DDI), protein-protein interaction (PPI) and clinical relation. DDI detection is an significant area of patient safety studies as these interactions can become very hazardous and boost the cost of health care. 
Similarly, the knowledge about interaction among proteins can help in understanding biological processes, complex diseases, and guiding drug discoveries \cite{kann2007protein}. In the clinical domain, the ability to recognize relations among medical concepts (treatments, tests, or problems) allows the automatic processing of clinical texts, resulting in clinical decision-making, clinical trial screening, and pharmacovigilance.
\\
In the vast literature on relation extraction, several techniques have been proposed to solve the problem ranging from the semantic-injected kernel model \cite{bunescu2005shortest, airola2008graph} to the machine learning embedded models \cite{thomas2013wbi,liu2015dependency}. In recent past, with the success of deep neural networks, the state-of-the-art models for these tasks have been drifted towards the deep learning frameworks \cite{choi2016extraction,zhao2016protein}. \\
\indent However, there have been a very few attempts in improving the performance of RE system irrespective of the tasks or domains.
One potential solution is to model the relation extraction problem in the multi-task learning framework, where a problem together with the other related problem can be learned by leveraging the shared representation. This method of multi-task learning provides advantages in \textbf{(1)} minimizing the number of parameters and \textbf{(2)} reducing the risk of over-fitting. The aim of multi-task learning (MTL) is to enhance the system performance by integrating the other similar tasks \cite{yadav2018multi,yadav2019unified}. 
When tasks are common and, in particular, when training data is limited, MTL can contribute to better results than a model trained on a single dataset, allowing the learner to capitalize on the commonality among the tasks. This can assist the overall model as dataset can contain information which are complementary to address individual tasks more correctly when trained jointly \cite{Crichton2017}.

However, most of the existing methods on multi-task classification tend to divide the feature space of each task into shared and private, depending solely on hard-parameter or soft-parameter sharing.
As such, these task-shared features are prone to being contaminated with external noise and task-specific features that often lead the system to suffer from the redundancy of the feature. \\
\indent To combat the contamination issue of task-shared features, in this paper, we propose an adversarial multi-task relation extraction framework. This framework deals with  utilizing  concept of adversarial learning to inherently disjoint the task-specific and task-invariant feature spaces. In  adversarial learning \cite{goodfellow2014generative}, a model is trained to correctly classify both unmodified data and adversarial data through the regularization method.
The adversarial learning paradigm provides an assurance that task-shared feature space is not contaminated with task-specific features and  contains only task-invariant features. \\
\indent In our study, we use the bi-directional gated recurrent units (Bi-GRU) \cite{cho2014properties} as a learning algorithm which has the capability to learn the features by capturing the long dependency information. Bi-GRU, unlike other Recurrent Neural Networks (RNN), such as Long Short Term Memory (LSTM) is computationally less expensive \cite{hochreiter1997long}. 
GRU also has the gating mechanism similar to LSTM to control the information flow. But unlike LSTM, GRU has no memory unit and has an update and reset gate.


As such, when down-sampling operation is performed on the output of GRU, we extract the optimal features from the entire input sequence covering the complete context information.  In the literature, the attention mechanism has shown the promising results in relation extraction by generating the optimal and effective features. In attention mechanism, a simple strategy has been followed by computing a weight vector corresponding to each hidden state of the RNN or CNN. The final hidden states are computed by performing the pooling operation (max, min, average etc.) on the weighted representations of hidden states. However, the computed attention weight focuses on a specific aspect of the input sequence. In this work, we attempt to capture multiple aspects of the input sequence by exploiting the self-attention mechanism. Basically, we learn to generate the multiple attention weight vectors, which eventually generate the multiple final representations of hidden states considering the various aspects of the input sequence.
\indent We apply our proposed approach on four popular benchmark datasets namely, AiMED \& BioInfer for PPI \cite{pyysalo2008comparative}, DrugDDI for DDI \cite{segura2013semeval}, and 2010 i2b2/VA NLP challenge dataset for clinical relation extraction (MCR) \cite{uzuner20112010}.  
Our proposed MTL model obtained the F1-Score points of $77.33$, $76.33$, $72.57$, and $81.65$ for AiMed, BioInfer, DDI and i2b2 relation extraction dataset, respectively.
We observe an average 5\% improvement in F-score in comparison to single task learning baseline model and over 3\% improvement on MTL baseline model.
Performance on any dataset does not decrease considerably, and performance increases significantly for all the four datasets. These are promising outcomes that set the potential for using the MTL model to solve the issue of biomedical RE. 
In addition to the baselines, our proposed model outperforms the state-of-the-art methods for all the datasets. This shows that, when we have tasks in common, multi-task can assist over the single-task models. The contributions of our proposed work can be summarized as follows:
\begin{enumerate}
\item We propose a multi-task learning (MTL) framework for relation extraction that exploits the capabilities of adversarial learning to learn the shared complementary features across the multiple biomedical and clinical datasets. We also exploit the self-attention mechanism which allows the final feature representation to directly access previous Bi-GRU hidden states via the attention summation. Therefore, the Bi-GRU does not need to carry the information from each time step 
towards its last hidden state.
\item Our proposed model is capable of automatically extracting the various relations (such as \textit{Protein-Protein Interaction}, \textit{Drug- Drug interaction: `int', `advice', mechanism, effect}, and relation between \textit{medical problem and treatment}, \textit{test and treatment}, \textit{treatment and treatment}). 
\item  We validate our proposed framework on four popular benchmark datasets (AiMED, BioInfer, SemEval 2013 Drug Interaction task, and i2b2 medical relation shared task dataset) for relation extraction, having different annotation schemes. 
\item Our unified multi-task model achieves the state-of-the-art performance and outperforms the strong baseline models for all the tasks in the respective datasets.
\end{enumerate}

\section{Related Works}
There has been recent surge in the interest of the BioNLP community to automatically detect and extract information from the scientific literature and clinical records \cite{savery2020chemical,goodwin2020customizable,srivastava2016recurrent,yadav2019information,ekbal2016deep,yadav2020exploring}.
In the past decade, there has been tremendous amount of the work on varieties of the relation extraction task such as extracting relationship between the bio-entities (proteins, gene, diseases, etc.) from the biomedical literature. Much previous works are done by using the Kernel-based technique which allows the representation learning of the data in the form of dependency structures and syntactic parse trees. Some of the other prominent techniques for extracting the relationships are based on the pattern-matching technique. Recently, with the success of deep learning technique, several techniques based on the Convolutional Neural Network, Recurrent Neural Network, and Long Short Term Memory network are widely utilized for extracting the relationships from biomedical literature and clinical records. Based on the tasks, we divide the related works in the following three categories:
\begin{itemize}
\item \textbf{Protein Protein Interaction task:} Several NLP techniques have been proposed to identify the relationships between the protein entities. The preliminary studies \cite{blaschke1999automatic,ono2001automated,bunescu2005comparative} on this task were essentially solved by using pattern-based model, where patterns were extracted from the data based on their syntactic and lexical properties. The main drawback with this approach is the inability to properly handle the complex relationship expressed in coordinating and relational clause. Dependency based approaches \cite{miyao2008evaluating,daraselia2004extracting} are more syntax aware techniques and have broader coverage than naive pattern based approaches. Some of the studies \cite{erkan2007semi} exploring the dependency based techniques incorporate the dependency information as a shortest dependency path between the sentences.  
Also the technique based on kernel method is often explored in the area of PPI extraction. Some of the prominent kernel-based approaches for extracting the PPI include edit-distance kernel \cite{erkan2007semi}, bag-of-word kernel \cite{saetre2007syntactic}, all-path kernel \cite{airola2008graph}, graph kernel \cite{airola2008all}, and tree-kernel \cite{zhang2006composite}. 
\cite{kim2010walk} proposed a walk-weighted sub sequence kernel that captures the syntactic structure by matching the e-walk and v-walk on the shortest dependency path. \cite{chang2016pipe} proposed a technique based on convolutional tree kernel by integrating the patterns of protein interaction. \\
\indent Recently, various studies have exploited deep learning based techniques \cite{yadav2018feature1,hsieh2017identifying} which do not require the manual feature engineering unlike the previous techniques based on the kernel, pattern and dependency based methods. \cite{hua2016shortest} first proposed the deep learning technique for extracting the relationship between the protein pairs. They used the CNN as a base learner over the word embedding generated through the Google News corpus. \cite{choi2018extraction} proposed a neural network framework which integrates several lexical, semantic and syntactic level information in the CNN model. Their study shows that integrating these additional information provides very minor improvement overall. \cite{peng2017deep} in their study proposed a two channel CNN technique for high level of feature extraction. In the first channel, they used words with additional syntactic features like part-of-speech, chunking information, dependency information, named entities and the word position information w.r.t the protein entities. In the second channel, they used the parent word information for every word. \cite{zhao2016protein} propose a greedy layer-wise unsupervised technique to extract the PPIs. They utilized the auto encoder on the unlabelled data for the parameter initialization of the deep neural network model and applied the gradient decent method using back propagation to train the whole network. \\
Various studies \cite{hsieh2017identifying} on PPI extraction task have also explored the Recurrent Neural Network framework. \cite{yadav2019feature} proposed a method based on the Bi-directional Long Short Term Memory Network (Bi-LSTM) equipped with the stacked attention mechanism. The input to their model is the shortest dependency path between the entity pairs. Their study shows that  providing multiple attentions can assist the model in better capturing the long contextual and structural information. \cite{ahmed2019identifying} proposed a tree RNN with structured attention framework for extracting the PPI information.

\item \textbf{Drug Drug Interaction task:}
Existing techniques on drug drug interaction can be categorized into one-stage and two-stage classification scheme \cite{SAHU201815}. In the one-stage classification \cite{bobic2013scai,hailu2013ucolorado_som}, the aim of the task is to identify the multiple relationships between the drug pairs, which could be from any of the interacting class or negative class. Several methods have explored the multi-class classifier, to capture the relationship between two target drugs in a sentence. While, in the two-stage classification \cite{bjorne2013uturku,rastegar2013uwm} scheme, there are two steps. The first step determines whether the target drug pair is interacting/non-interacting. In the second step, only the interacting sentences are considered as  inputs to the multi-class classifier. These approaches can further be classified into handcrafted feature and latent feature based methods. Techniques \cite{kim2015extracting,bokharaeian2013nil_ucm,chowdhury2013fbk} based on the hand-crafted feature mainly utilize support vector machines (SVMs) as the base learner. These techniques are reliant on several hand-crafted features such as Part-of-Speech, chunk, syntax trees, dependency parsing, and trigger words. These methods are utilized in other biomedical relation extraction tasks, such as adverse drug reactions extraction tasks \cite{harpaz2014text,xu2015large}, protein protein interaction extraction \cite{qian2012tree}, relation extraction between diseases and genes \cite{bravo2015extraction}, and relations between the medical concepts \cite{rink2011automatic}. These techniques have appeared to perform well, however they are very domain-specific and are dependent on other NLP tools.  
Approaches exploring the latent features, are based on deep learning models, that are proved to be powerful solutions to the feature based models. 
Below we provide the survey in detail for the above described methods:
\begin{itemize}
    \item \textbf{Linear Methods:} This method utilizes a linear classifier that takes as an input the domain specific or manually designed features. The system proposed by Uturku \cite{bjorne2013uturku} explored Turku event extraction system (TEES) for identifying the drug interaction pairs. TEES utilizes the features of dependency parsing and lexicon derived from MetaMap and Drugbank. \cite{rastegar2013uwm} developed two-stage classification technique based on SVM. They explored several hand-crafted features such as lexical, contextual, semantics and tree structured features.
    \item \textbf{Kernel Methods:} These techniques are more advanced than linear methods, where they have explored graph based features. 
All-paths graph kernel \cite{Airola2008} learns a classifier based on the weighting scheme for dependency parse tree feature and surface feature. k-band shortest path spectrum \cite{tikk2010comprehensive} utilizes the shortest dependency path between the entity pairs to build a classifier. It further permits the mismatches for variables and includes all the nodes within k-distance from the shortest path. The shallow linguistic kernel \cite{giuliano2006exploiting}, as the name suggests, captures the shallow linguistic features such as part-of-speech, stem, word and other morphological features that also explore the properties of the surrounding words. 
\cite{thomas2011relation} proposed an ensemble based DDI extraction system. They explored the various kernel classifiers in addition to the case-based reasoning technique for classifying the drug pairs. \cite{chowdhury2013fbk} also explored various kernel classifier by integrating multiple kernel methods such as SL kernel, mildly extended dependency tree (MEDT) kernel and path-enclosed tree (PET) kernel. PET kernel captures the smallest subtree involving the two entities in a phrase structure parse tree. MEDT kernel uses linguistically motivated expansions to capture the prominent clue words between the entity pairs. The system proposed by \cite{chowdhury2011two} also utilizes several kernels based on the feature and tree kernel methods. Precisely, they explored MEDT, SL, PET, global context and local context kernel.  
\item \textbf{Deep Learning Technique:} Deep learning technique is based on the neural network approach that utilizes latent features instead of hand-crafted features. This technique encodes the word level representation using neural network for the generation of sentence level features. For the final classification, the network uses the sentence level feature. In the recent past, several deep learning methods have been used to address DDI tasks. SCNN system was proposed by \cite{zhao2016drug}. They have used convolution neural network to capture more dense sentence representation feature. SCNN also utilized some additional features such as PoS and dependency tree based feature. \cite{SAHU201815} advanced the state-of-the-art technique by proposing three different models based on the concept of LSTM and attentive pooling. All of these models takes as an input the word and position embedding and does not rely on any hand-crafted features. Some of the other prominent work that has explored deep learning framework are \cite{zheng2017attention,kavuluru2017extracting,wang2017dependency,asada2017extracting,lim2018drug}
\end{itemize}

\item \textbf{Medical Concept Relation Extraction task:}
Electronic medical records such as patient's discharge summaries and progress notes contains information about the medical concepts and relationship. 
To aid an advanced patient care, it is required have a technique for automatic processing of the clinical records. To address this issue, Informatics for Integrating Biology \& the Bedside (i2b2) organized a shared task challenge \cite{uzuner20112010} that aims to identify the relationship between medical concepts i.e. problems, treatments, or tests from the EMR document. Identifying the correct relationship between the medical concepts requires the knowledge of the context in which two concepts are discussed.
Existing techniques for extracting the relations between the medical concepts can be grouped into semi-supervised and supervised classes. 
\cite{de2010nrc} explored the semi-supervised method to determine the relationship between a concepts-pairs. They used maximum entropy as an classification algorithm to train separately three classifier for each concept pair  i.e. test-problem, treatment-problem, and problem-problem relations. They explored various external features obtained from other NLP pipeline such as cTAKES and MetaMap. Additionally, they also used word-level feature, PoS tags, dependency path based feature, and distance feature that capture minimal, average and maximal tree distances to the common ancestor. To overcome the label imbalance problem in the training data, they computed the relatedness between two medical concepts by using pointwise mutual information in MEDLINE and bootstrapping with unlabeled examples.\\

\begin{figure*}[t]
\centering
\includegraphics[width=10cm,height=8cm]{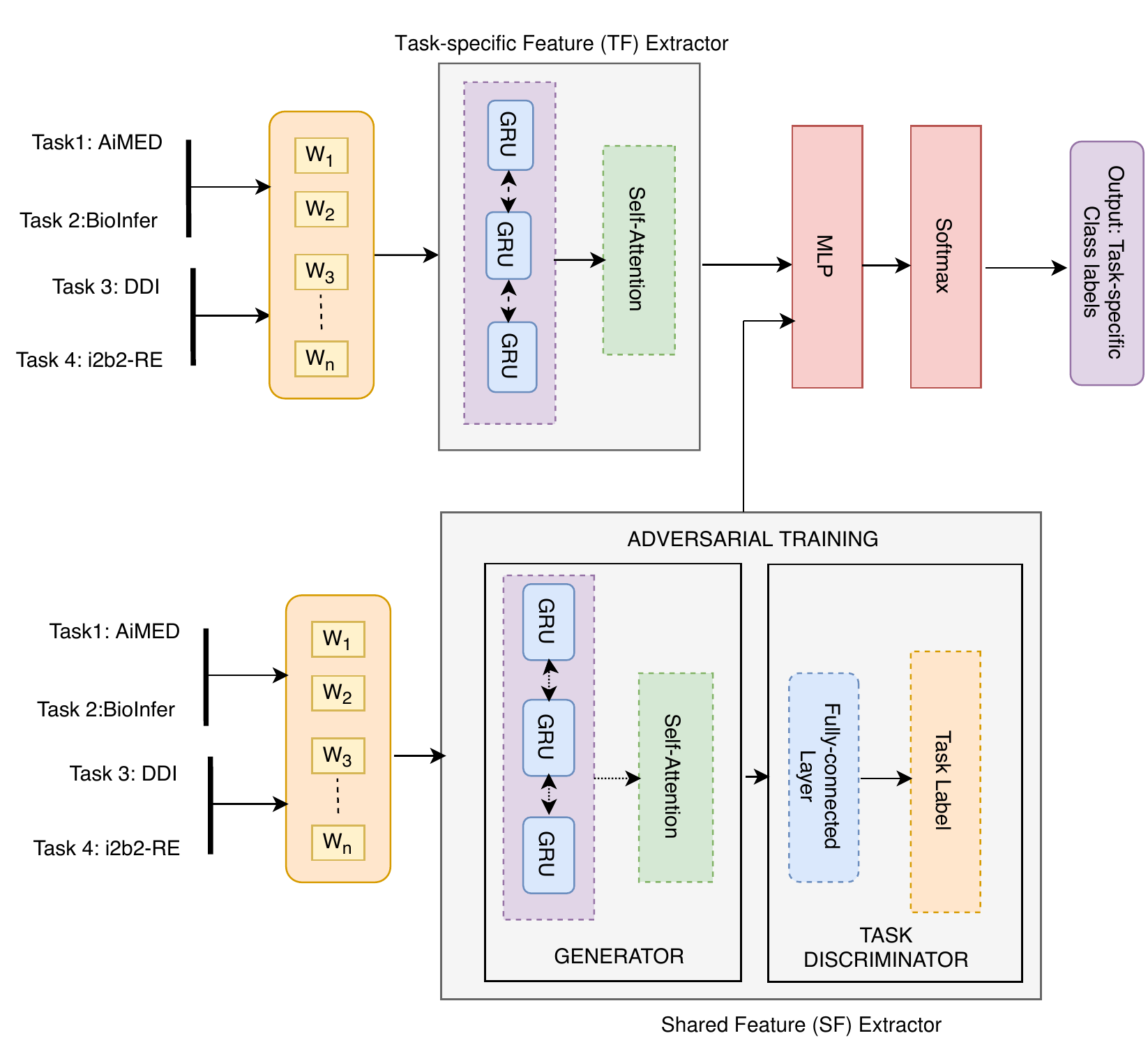}
\label{figure-proposed}
\caption{Proposed multi-task model for various biomedical relation extraction tasks.}
\end{figure*}
Majority of the research work has been carried out on this problem are highly dependent on the supervised approach. In those works \cite{kang2010erasmus,jiang2010hybrid,grouin:hal-00795663}, statistical machine learning classifier (CRF, SVM) is used to identify the relations. As the i2b2 shared task dataset, contained a large portion of concept pairs without any relations, some of the system \cite{roberts2010extraction,jonnalagadda2012enhancing,de2011machine} proposed two stage classification methods, where in the first stage, the concept pair with relation/no relation were classified. In the latter stage, only those concept pair with the relation participate for identifying the given relationship. All of the participating system heavily rely on the hand-crafted features. They exploited semantic, lexical, syntactic, domain-specific ontology feature. The system proposed by \cite{demner2010nlm} used medical knowledge graph (UMLS) concept identifiers and applied feature selection technique to capture more relevant feature. \cite{grouin:hal-00795663} used linguistic feature to complement their machine learning component to extract the medical concept relations.  
Recently, neural network techniques are widely adopted for the clinical relation extraction task. \cite{sahu2016relation} explored the capabilities of convolutional neural network to capture prominent feature for extracting relations. Another study conducted by \cite{LUO201785} used two variant of RNN-LSTMs network, segment level LSTM and sentence level LSTM for encoding the relation. The experimental results shows that the proposed approach perform comparable w.r.t state-of-the-art system. They also identified that word embedding from clinical text is more beneficial than the general text.\\
Other prominent study was conducted by \cite{rosario2004classifying} to identify the relations among disease and treatment. They used several neural network and graphical models. Furthermore, they utilized other hand-crafted features such as lexical, semantic, and syntactic feature for classification. They conducted their study on very small dataset consisting of biomedical research article. \cite{bundschus2008extraction} applied CRF for extracting relations between disease, treatment and gene. They proposed two step model where in the first step, they identified the entities and in the second step, they extract the relationship. For both the steps, they explored CRF as the base learner. \cite{bravo2015extraction} developed a system for finding the association between disease, drug and target in EU-ADR dataset. They exploited kernel based method that uses the shallow linguistic and dependency kernel for extracting the relations. \cite{ningthoujam2019relation} developed shortest dependency path based deep neural network framework that also utilizes other features like Parts-of-speech information, dependency labels, and the types of the entities.
\end{itemize}
\section{Materials and methods}\label{material_methods}
In this section, we describe the methodology used to extract the relations from various biomedical texts in our proposed multi-task learning framework. We begin by the problem statement followed by the introduction of the Gated Recurrent Unit (GRU) which is used in our models as a base learning algorithm.  
Following that, we describe the proposed multi-task model in detail that utilizes the concept of adversarial learning. \\
\noindent \textbf{Problem Statement:}\\
Given an input text sequence $S$ consisting of $n$ words, i.e., $S = \{w_1, w_2 \ldots w_n\}$ and a pair of entities \footnote{The entity pair can be the \textit{protein}, \textit{disease}, \textit{problem} name, etc. } $(e_1, e_2)$ where $e_1 \in S$ and $e_2 \in S$.
The task is to predict the maximum probable class $\bar{y}$ from the set of class labels, $Y$. Mathematically, 
\begin{equation}
    \bar{y}=\argmax_{y \in Y} prob(y |S, e_1, e_2, \theta)
\end{equation}
where, $\theta$ is the model parameter. Each token $w_i \in S $ of the input sequence $S=\{w_1, w_2, \ldots, w_n\}$ is mapped into $d$ dimensional word embedding sequence $x=\{x_1, x_2, \ldots, x_n\}$, where $x_i \in \mathcal{R}^{d}$.
\subsection{Gated Recurrent Units}\label{GRU}
\label{bi-gru}
In this work, we use GRU\cite{cho2014properties} as the base learner. The GRU is an improved version of the recurrent neural network and being used as variant of the LSTM due to its simpler architecture over the LSTM. Similar to the other RNN, GRU has the capability of internal memory. This internal memory helps them to exhibit temporal dynamic behaviour for a time sequence. Additionally, GRU is able to solve the problem of the vanishing gradient problem which comes with a standard recurrent neural network. Similar to LSTM units, GRU also have the gating mechanism to control the flow of information and produce the effective hidden state representation.  

GRU has two neural gates, \textit{update} and \textit{reset} gate. The task of \textit{update} gate is to helps the model for determining the amount of information need to be carry forward along to the future. The \textit{reset} gate helps the model to determine the amount of past information need to be forget. Specifically, a GRU network successively reads the input token $x_{i}$, as well as the visible state $h_{i-1}$, and generates the new states $c_{i}$ and $h_{i}$.
\begin{equation}
\begin{split}
{z}_{i}&=\sigma (\mathbf{W}^{z}x_{i}+\mathbf{V}^{z}{h}_{i-1} + \mathbf{b}^{z} )  \\
{r}_{i}&=\sigma (\mathbf{W}^{r}x_{i}+\mathbf{V}^{r}{h}_{i-1} + \mathbf{b}^{r} )  \\
{c}_{i} &=tanh (\mathbf{W}x_{i}+\mathbf{V} ({r}_{i} \odot {h}_{i-1}) + \mathbf{b})\\
{h_i}&=z_{i} \odot {c}_{i} +  (1- {z}_{i}) \odot {h}_{i-1}
\end{split}
\end{equation}
where ${z}$ and ${r}$ are the update and reset gates, respectively. The final representation at a given time $t$, from a bi-directional GRU (Bi-GRU), can be computed by concatenating the forward $\overrightarrow{h_t}$ and backward $\overleftarrow{h_t}$ hidden states. From here onward, we will call \texttt{Bi-GRU} as function having the inputs, $x_t$ and $h_{t-1}$ and output $h_t$.
\subsection{Relation Extraction Framework}
In the relation extraction framework \textbf{(c.f. Figure-1)} the input sequence $x=\{x_1, x_2, \ldots, x_n\}$ with the corresponding entity pair $(e_1, e_2)$ is transformed into the hidden state representation using the Bi-GRU (c.f. section \ref{bi-gru}). In order to emphasize the given entity pair, the corresponding entity word in the input sequence is marked with the special token $ENTITY$ and assigned a fixed word embedding to it. More formally, the hidden state at each time step is calculated as follows:
\begin{equation}
\label{stl-eq}
h_t = \text{Bi-GRU}(x_t, h_{t-1})
\end{equation}
Let the hidden state dimension for each Bi-GRU unit be $d_h$. We formulate a hidden state matrix $H \in \mathbb{R}^{n \times d_h}$.
\begin{equation}
    H=\left( \mathbf{h_1}, \mathbf{h_2},\cdots \mathbf{h_n} \right)
\end{equation}
We compute the effective input sequence encoding $\bar{h}$
using function\footnote{This function is adversarial learning based self-attentive network which computes the effective input sequence encoding.} $fun(\bar{h}; \theta)$ with learning parameter $\theta$. The input sequence encoding $\bar{h}$  is fed to a fully connected softmax layer to generate the probability distribution over the predefined classes.
\begin{equation}
\label{softmax}
\bar{y} = softmax(\bar{h}^T\mathbf{W}+\mathbf{z})
\end{equation}
Here, $\mathbf{W}$ and $\mathbf{z}$ are weight matrix and bias vector, respectively. The term $\bar{y}$ denotes the predicted probability distribution.
\indent Let us have a training dataset with $N$ samples such that $\{(x_1, y_1), (x_2, y_2), \ldots, (x_n, y_n)\}$. The network parameters are trained to minimize the loss function-- cross entropy of the probability distributions of predicted ($\bar{y}$) and 
true class ($y$) over the $C$ number of classes.
\begin{equation}
\label{singleLoss}
\mathcal{L_{CE}}(\bar{y}, y)=-\sum_{i=1}^{n}\sum_{j=1}^{C}y_i^j log(\bar{y}_i^j)
\end{equation}


\subsubsection{Attentive Mechanism: }\label{attention-1}
We encode the input sequence with by adopting the self-attentive mechanism\cite{lin2017structured} over the Bi-GRU generated hidden state sequence.
The input to the attention mechanism is the Bi-GRU hidden states $H$. The self-attention generate the attention weight vector $v$ computed as follows:

\begin{equation}
\label{ca-attention}
\begin{aligned}
     \mathbf{p} &=  tanh \left({ U }H^T\right) \\
     \mathbf{v} &= softmax \left( \mathbf{w} p \right)
\end{aligned}
\end{equation}
Here $U \in \mathbb{R}^{d_a \times d_h}$, and $\mathbf{w} \in  \mathbb{R}^{d_a}$ and $d_a$ is a hyperparameter and $d_h$ is the size of hidden state. The final hidden state representation $s$ is computed by the weighted (weight provided by $\mathbf{v}$) sum of the each time step of the Bi-GRU. The major drawbacks of the aforementioned representation is that it focuses on the specific component of the input sequence. The specific component could be a relation between a given entity to other words in the sequence. We call it is a aspect to represent the input sentence. In order to capture the multiple aspects, we required multiple $\mathbf{m}$'s that capture the various notion of the input sequence. Therefore, we extend the attention mechanism from focusing on single aspect to multiple aspects. Let us assume, we want to extract `$a$' number of multiple aspects from the input sequence. To achieve this, we extend the Eq \ref{ca-attention} as follows:
\begin{equation}
\label{cam-attention}
\begin{aligned}
     \mathbf{p} &=  tanh \left({ U }H^T\right) \\
     \mathbf{V} &= softmax \left( \mathbf{W} p \right)
\end{aligned}
\end{equation}
Here, $\mathbf{W} \in \mathbb{R}^{a  \times d_a}$. Formally,
We compute the `$a$' weighted sums by multiplying the  matrix $V$ and LSTM hidden states $H$. We obtain a matrix representation $M$ of the input sequence by multiplying the attention weight matrix $V$ and hidden state representation $H$.
\begin{equation}
\label{attention-equation}
    M=VH
\end{equation}
We concatenate each row of the matrix representation to get the final vector representation of the input sequence.
\subsection{Multi-task Learning for Relation Extraction}\label{shared-private}
In this study, we introduce the novel method for biomedical relation extraction exploiting the adversarial learning in the multi-task deep learning framework. Our model leverages joint modeling of the entities and relations in a single model by exploiting attentive Bi-GRU based recurrent architecture. We propose an adversarial multi-task learning with attention (Ad-MTL) model for relation extraction task. 
Multi-task learning exploits the correlation present among similar tasks to improve classification by learning the common features of multiple tasks simultaneously. We build a latent feature space that holds the features that are common to various tasks. Specifically in the model, the outputs generated at each time-step of the shared Bi-GRU, are considered to be common latent features. We generate the task-specific feature features for each task by task-specific Bi-GRU network equipped with the self-attentive network discussed in section \ref{attention-1}.

We compute the two hidden states at each time step $t$ for a given task $k$, one task-specific hidden state, $h_t^k$, and another shared hidden state, $s_t^k$. The former captures the task dependent features and the latter captures the task invariant features. Both the hidden state representations are computed similar to Eqn. \ref{stl-eq}.
\begin{align}
s_t^k &= \text{BI-GRU}(x_t^k,s_{t-1}^k, \theta_s) \\
h_t^k &= \text{BI-GRU}(x_t^k,h_{t-1}^k, \theta_h)
\end{align}
where $\theta_s$ and $\theta_h$ are Bi-GRU's parameters, $x_t^k$ denotes the input at time $t$. We generate the task-specific feature representation by applying self-attention using equation \ref{attention-equation}. We also use a feed-forward network with a hidden layer to project the attentive feature representation to another vector space. We call the final task-specific feature representation as of task $k$ as $TF_k$. 
Similar to the task-specific feature generation, we use a feed-forward network to project the the shared feature into another vector space and call it is a shared feature $SF$.
The concatenation of the shared and task specific features is fed into a fully connected layer followed by the softmax layer. The softmax layer returns the class distributions, $y_{pred}^k$, for the underlying task, $k$.
For every task, $k$, with training samples ($x_i^k$,$y_i^k$), both the task specific parameters and shared parameters are optimized to minimize the cross-entropy of the predicted ($\hat{y}_i^k$) and actual probability distributions ($y_i^k$), whose loss is computed as:
\begin{equation} \label{shareCrossEnt}
\mathcal{L}_{CE}^k =   \mathcal{L}_{CE}(\hat{y}_i^k , y_i^k)
\end{equation}
\noindent
where $ \mathcal{L}_{CE}(\hat{y},y)$ is defined as Eq. \eqref{singleLoss}. 

\subsubsection{Adversarial Training}
The above discussed model though intended to separately host the shared and task-specific features, provides no guarantee to behave so, there might be contamination of shared features in the task-specific feature space and vice versa. To handle this, we exploited the principle that a good shared feature space has features that make it impossible to predict the source task of the feature.
For achieving the above, a \texttt{Task Discriminator} $D$ is used to map the attention prioritized shared feature to estimate the task of its origin. In our case, \texttt{Task Discriminator} is a fully connected layer using a softmax layer to produce the probability distribution of the shared features belonging to any task. A Bi-GRU works as \texttt{Generator} $(G)$ to generate shared features. This Bi-GRU layer is made to work in an adversarial way, preventing the discriminator from predicting the task and hence preventing contamination in the shared space. The adversarial loss is used to train the model. Similar to \cite{goodfellow2014generative,liu-qiu-huang:2017:Long}, we use the following adversarial loss function
\begin{equation}
\label{advloss}
\mathcal{L}_{adv}=\min_{G}\Big(\max_D \big( \sum_{t=1}^T\sum_{i=1}^{N_t} d_i^tlog\big[D\big(G(x^t))]\big)\Big)
\end{equation}
where $d_i^t$ is the gold label indicating the type of the current task. The min-max optimization problem is addressed by the gradient reversal layer \cite{ganin2014unsupervised}. The total loss of the network will be as follows:
\begin{equation}
\label{total-loss}
    \mathcal{L}_{total}=\alpha \sum_{k=1}^{K} \mathcal{L}_{CE}^{k} + \beta \mathcal{L}_{adv} 
\end{equation}
where $\alpha$ and $\beta$ are the scalar parameters.

\section{Experimental Results and Discussion}
In this section, we will begin by briefly describing the various tasks and the datasets followed by the experimental results and analysis.
In this study, we focus on the following tasks:
\subsection{Protein Protein Interaction Extraction (PPI):}The goal of this task is to classify 
whether or not a sentence containing a proteins pair actually indicates interaction between the pair. Here, we considered the positive instances as the interacted protein-pair and the negative instances as the non-interacted protein pair. If the relationship between the protein pair is not explicitly provided, the pairs were considered to be the negative instances. In order to identify those instances, we extracted all possible proteins pairs from the sentences. \\
We utilize two standard benchmark datasets for PPI tasks, namely AiMed and BioInfer\footnote{http://corpora.informatik.hu-berlin.de/}. AiMed dataset is derived from the $197$ abstracts of the Database of Interacting Proteins (DIP) and contains manually tagged relationship between the protein entities. There are in total $5834$ relationship out of which $1000$ are interacting relation and $4834$ non-interacting relation. \\
BioInfer (Bio Information Extraction Resource) is another manually annotated PPI dataset developed by Turku BioNLP group\footnote{http://bionlp.utu.fi/} that contains over $1100$ sentences. It has $2534$ instances of protein interacting relationship and $7132$ non-interacting relationship.



\subsection{Clinical Relation Extraction (MRE): }\label{ssec:cre} This task aims to extract the relation between the clinical entities (\textit{Problem, Treatment,and Test}) from the EMR. For this, we used the benchmark dataset released by i2b2 as a part of i2b2 2010 clinical information challenge \cite{uzuner20112010}. 
The dataset was collected from three different hospitals, which consists of discharge-summaries and progress notes of the patients those were manually annotated by the medical practitioners for identifying
the three major relation types: medical problem–treatment (TrP) relations, medical problem–test (TeP) relations, and medical problem–medical problem (PP) relations. These relations were further fine-grained into $8$ different relation types which were: treatment caused medical problems (TrCP), treatment administered medical problem (TrAP), treatment worsen medical problem (TrWP), treatment improve or cure medical problem (TrIP), treatment was not administered because of medical problem (TrNAP), test reveal medical problem (TeRP), Test  conducted  to  investigate  medical problem (TeCP), and Medical problem indicates medical problems (PIP). The exact definition of each of these relation types can be found in \cite{uzuner20112010}. \footnote{While, the actual dataset released during the challenge was having $394$ documents for training and $477$ documents for testing. However, we were able to download only $170$ documents for training and $256$ documents for testing as also pointed out by \cite{sahu2016relation} from i2b2 website.}\\
It is to be noted that since we did not have enough  training  samples  for  all  relation  classes present in the dataset, we have removed following three classes:  TrWP, TrIP, and TrNAP. this same strategy was also followed by the \cite{sahu2016relation}.

\subsection{Drug Drug Interaction Extraction (DDI):} \label{ssec:ddi}  
Given a sentence with two pharmacological substances, this task aims to classify if the given drug pair interacts with each other or not. For this task, we utilize the standard benchmark DDI corpus from Semeval 2013 DDIExtraction challenge dataset \cite{segura2013semeval}. The DDIExtraction 2013 task exploits the DDI corpus, which contains MedLine abstracts on drug-drug interactions as well as documents describing drug-drug interactions from the DrugBank database. The corpus consists of total 1,017 abstracts from Medline (233) and DrugBank (784) databases which were manually annotated to obtain 18,491 pharmacological substances and 5,021 drug-drug interactions. Here each interacted drug pair is further classified into one of four types, namely mechanism, advice, effect, and int. It is to be noted that during the challenge, original dataset has 23756 false samples for training and 4737 for testing. But when we obtained the data set from shared task organizers, there were only 22474 false training instances and 4461 testing sample for false class. In the Table-\ref{ddi2013}, we have provided the statistics of the DDI dataset. 

\begin{figure*}[t]
\centering
\includegraphics[width=10cm,height=8cm]{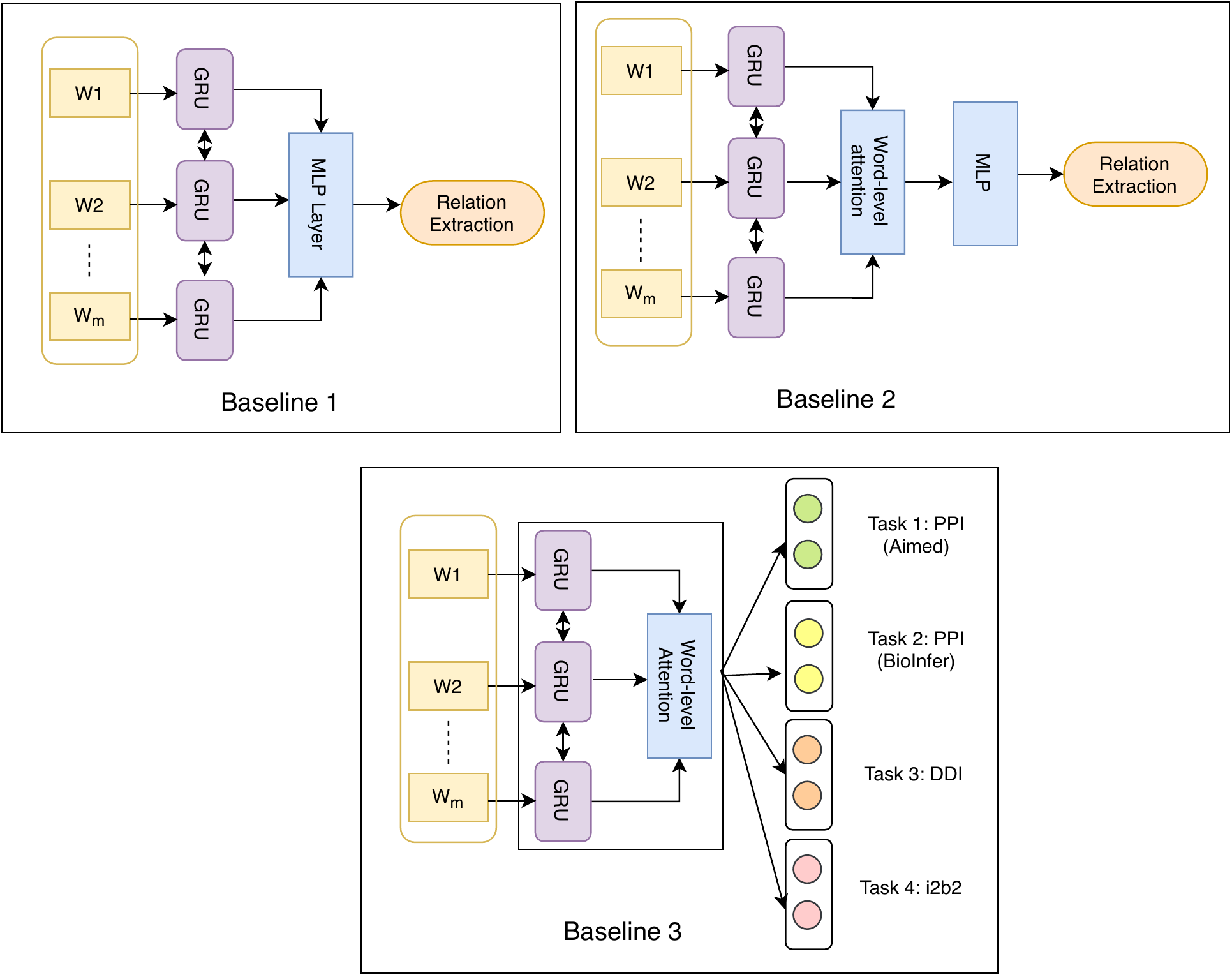}
\label{figure-baseline}
\caption{Baseline models for various biomedical relation extraction tasks.}
\end{figure*}
The detailed statistics of all the three datasets are reported in the Table-\ref{binary datasets},\ref{i2b2relxtr},\ref{ddi2013}
\begin{table}[h]
\centering
      \begin{tabular}{ccccc}
        \hline 
        Dataset	 & Interacting Pairs   & Non-interacting pairs   \\ \hline \\
		AiMed PPI &1000  & 4834 \\
		BinInfer PPI  & 2534 & 7132 \\  \hline 
         \end{tabular}
     \caption{Statistics: AiMed and BioInfer Dataset}
\label{binary datasets}
\end{table}
\begin{table}[h]
\centering
      \begin{tabular}{ccc}
        \hline
        Label  & Relation  & No.of Samples\\
        \hline 
        0 & TrIP &203 \\
        1 & TeRP & 3053 \\
        2 & TrAP& 2617 \\
        3 & PIP & 2203 \\
        4 & TeCP & 504 \\
        5 & TrCP & 526 \\
        6 & TrNAP & 174 \\
        7 & TrWP & 133 \\
        8 & NONE & 54530\\
        \hline
      \end{tabular}
      \caption{Statistics: 2010 i2b2/VA NLP Challenge dataset}
\label{i2b2relxtr}
\end{table}
\begin{table}[h]
\centering
      \begin{tabular}{cccc}
        \hline
        Label  & Relation  & No.of Samples (Train) & No.of Samples (Test) \\
        \hline 
        0 & False &22474 & 4461 \\
        1 & effect & 1685 &360 \\
        2 & mechanism & 1316 &302 \\
        3 & advice & 826 &221 \\
        4 & int & 188 &96 \\
        \hline
      \end{tabular}
       \caption{Statistics: Semeval 2013 DDIExtraction challenge dataset}
\label{ddi2013}
\end{table}

\begin{table*}[]
\centering
\resizebox{\textwidth}{!}{%
\begin{tabular}{l|c|c|c|c|c|c|c|c|c|c|c|c}
\hline
\multirow{2}{*}{\textbf{Tasks}} & \multicolumn{3}{c|}{\textbf{Baseline 1}} & \multicolumn{3}{c|}{\textbf{Baseline 2}}  & \multicolumn{3}{c|}{\textbf{Baseline 3}} & \multicolumn{3}{l}{\textbf{Proposed Approach (MTL)}}  \\ \cline{2-13} 
 & P & R & F & P & R & F & P & R & F & P & R & F \\ \hline
\textbf{PPI-AiMED} &  69.61  & 71.16  & 70.37  & 67.75 &  69.65  & 68.69 & 75.85 & 72.74  & 74.27  &78.12 & 76.56 & \textbf{77.33 }  \\ \hline
\textbf{PPI-BioInfer}  & 71.29  &70.95  &71.12  & 68.47 & 69.03  & 68.75 & 75.44 & 73.79 & 74.61 & 76.69 & 75.98 &\textbf{ 76.33 } \\ \hline
\textbf{DDI} &  71.99 & 67.26 & 67.46 & 70.24  & 65.15  & 67.60 & 75.82 & 64.50 & 69.71 & 76.52 & 69.01 & \textbf{72.57}  \\ \hline
\textbf{i2b2-2010 Clinical relation} & 81.72 & 78.52  & 80.09  & 82.28  & 78.96  & 80.59 & 81.06 & 80.36 & 80.71 & 81.92 & 81.37 &\textbf{ 81.65}  \\ \hline
\end{tabular}
}
\caption{Evaluation results of proposed MTL model and baselines system. Performance is reported in terms of `P': Precision, `R': Recall, and `F': macro-F1-Score. Baseline 1 is single task learning model based on Bi-GRU. Baseline 2 is STL model with Bi-GRU + word-level attention. Baseline 3 is MTL model with a shared Bi-GRU layer. All the results are statistically significant as \textit{p-value} $<$ $0.05$). }
\label{tab:eval-proposed}
\end{table*}
\begin{table}[h]
      \begin{tabular}{cc}
        \hline 
        Hyper-parameters & \thead{Value}   \\ \hline \\
        Max sentence length & 60 \\
        Embedding dimension & 200 \\
		GRU Hidden State Dimension & 64 \\
		Attention Size ($d_z$) & 350 \\
		Attention aspect Size ($a$) & 5 \\
        \# of hidden neuron in Feed-forward n/w  & 100 \\  
    Activation & \textit{relu} \\
    Dropout rate & 0.3 \\
        Output Activation &  Soft-max  \\
        Optimizer & Adam Optimizer \\
        Learning Rate & 0.001 \\
        $\alpha$ & 1 \\
    $\beta $ &  0.05 \\
        \hline  
      \end{tabular}
      \caption{Optimal hyper-paramete values on  proposed model}
\label{hyperparams}
\end{table}

\section{Network Training and Hyper-parameters Setting}
We train the network by minimizing the total loss of the network (Eq. \ref{total-loss}).
In adversarial training, first we pre-trained the discriminator to avoid the instability in the network. To pre-train the discriminator, we use a Bi-GRU network to get the representation of the sentences from the different tasks. We have shown the training process in Algorithm \ref{algo:mt}. The shared feature extractor model $S$ in Algorithm \ref{algo:mt} is a Bi-GRU network followed by self-attention layer and it is being exploited by all the tasks (c.f. Section \ref{shared-private}). In the
shared feature extractor, there is an additional adversarial learning component, where feature extractor (Generator) operates adversarially towards a learnable multi-layer perceptron (Discriminator), preventing it from making an
accurate prediction about the types of the task the
features generated from.  For generating the word embedding, we have used pre-trained embedding available at \footnote{http://evexdb.org/pmresources/vec-space-models/}.
\begin{table*}[]
\resizebox{\textwidth}{!}{%
\begin{tabular}{c|ccc|ccc|ccc|ccc}
\hline
\multirow{2}{*}{\textbf{Model}} & \multicolumn{3}{c|}{\textbf{AiMED}} & \multicolumn{3}{c|}{\textbf{BioInfer}} & \multicolumn{3}{c|}{\textbf{DDI}} & \multicolumn{3}{c}{\textbf{i2b2-RE}} \\ \cline{2-13}
 & P & R & F1 & P & R & F1 & P & R & F1 & P & R & F1\\ \hline
\textbf{\begin{tabular}[c]{@{}c@{}}Proposed Model\\ Multi-task adversarial learning + GRU + self attention\end{tabular}} & 75.06 & 75.10 & 75.08 & 76.15 & 75.05 & 75.59 & 76.99 & 68.09 & 72.27 & 82.16 & 80.79 & 81.47 \\ \hline
\textbf{-self attention} & 71.13 & 71.63 & 71.05 & 73.63 & 72.21 & 72.66 & 75.09 & 64.98 & 68.53 & 79.69 & 80.49 & 79.90 \\ \hline
\textbf{-adversarial learning} &76.55  &74.10  &75.30  &75.92  &75.34  &75.63  &77.32  &68.28  &72.52  &81.65  &81.06  &81.35  \\ \hline
\end{tabular}%
}
\caption{Ablation study by jointly training all the four dataset together.}
\label{ablation}
\end{table*} 

\begin{table*}[]
\resizebox{\textwidth}{!}{%
\begin{tabular}{c|ccc|ccc|ccc|ccc}
\hline
\multirow{2}{*}{\textbf{Model}} & \multicolumn{3}{c|}{\textbf{AiMED}} & \multicolumn{3}{c|}{\textbf{BioInfer}} & \multicolumn{3}{c|}{\textbf{DDI}} & \multicolumn{3}{c}{\textbf{i2b2-RE}} \\ \cline{2-13}
& P & R & F1 & P & R & F1 & P & R & F1 & P & R & F1\\ \hline
\textbf{\begin{tabular}[c]{@{}c@{}}Proposed Model\\ Multi-task adversarial learning + GRU + self attention\end{tabular}} & 78.12 & 76.56 & 77.33 & 76.69 & 75.98 & 76.33 & 76.52 & 69.01 & 72.57 & 81.92 & 81.37 & 81.65 \\ \hline
\textbf{-self attention} & 72.96 & 73.21 & 73.09 & 74.91 & 73.98 & 74.44 & 75.34 & 64.92 & 69.74 & 80.74 & 79.16 & 79.94 \\ \hline
\textbf{-adversarial learning} &78.26  &75.43  &76.82  &76.57  &75.72  &76.14  &77.92  &67.31  &72.23  &83.68  &81.02  &82.33   \\ \hline
\end{tabular}%
}
\caption{Ablation  study  by training two similar dataset (AiMed + BioInfer) and (DDI+i2b2-RE) at a time.} 
\label{ablation-1}
\end{table*}

\begin{table*}[]
\centering
\resizebox{\textwidth}{!}{%
\begin{tabular}{llcccccc}
\hline
\multirow{2}{*}{\textbf{System}} & \multicolumn{1}{c}{\multirow{2}{*}{\textbf{Technique}}} & \multicolumn{3}{c}{\textbf{AiMED}} & \multicolumn{3}{c}{\textbf{BioInfer}} \\ \cline{3-8} 
 & \multicolumn{1}{c}{} & \textbf{P} & \textbf{R} & \textbf{F} & \textbf{P} & \textbf{R} & \textbf{F} \\ \hline
\textbf{Our System} & \textbf{MTL-Adversarial (Bi-GRU+ self-attention)} & \textbf{78.12} & \textbf{76.56}  & \textbf{77.33} & \textbf{76.69} & \textbf{75.98}  & \textbf{76.33}\\ \hline
\cite{hsieh2017identifying}$^{*}$ & LSTM$_{pre}^{*}$ & 73.12 & 73.02 & 73.07 & 73.81 & 72.69 & 73.24 \\
\cite{hua2016shortest} & sdpCNN (SDP+CNN) & 64.80 & 67.80 & 66.00 & 73.40 & 77.00 & 75.20 \\ 
\cite{qian2012tree} & Single kernel+ Multiple Parser+SVM & 59.10 & 57.60 & 58.10 & 63.61 & 61.24 & 62.40 \\ 
\cite{peng2017deep} & \begin{tabular}[c]{@{}l@{}}McDepCNN \end{tabular} &  67.3 & 60.1 & 63.5 & 62.7 &  68.2 & 65.3 \\
\cite{zhao2016protein} & \begin{tabular}[c]{@{}l@{}}Deep neural network\end{tabular} &  51.5 &  63.4 & 56.1 & 53.9 &  72.9 & 61.6 \\ \hline
\end{tabular}%
}
\caption{Comparison with the SOTA techniques for PPI task on AiMed and BioInfer datasets. Performance is reported in terms of `P': Precision, `R': Recall, and `F': F1-Score (macro). \cite{hsieh2017identifying}$^{*}$ denotes the re-implementation of the systems proposed in [32] with the authors reported experimental setups using their publicly available source codes.} 
\label{results}
\end{table*}

\subsection{Evaluation Criteria}
We evaluate the performance of all our models using  macro-averaged precision,  macro-averaged recall and  macro-averaged F1-Score metrics. Due to the unavailability of  separate validation set for AiMED, BioInfer, and i2b2-2010 clinical relation task datasets, we adopt $10$-fold document cross-validation strategy to compute the precision, recall, and F1-score values. We consider the predicted class-label as the correct only if it has the exact match with the ground truth annotation. For the DDI task, we report the performance of the models on the test set.
\begin{algorithm}[ht!]
 \SetAlgoLined
Set the max number of epoch: $epoch_{max}$.\\
\For{$t$ in $\{ t_1, t_2, \ldots, t_K \}$ }
{
    \textbf{1.} Pack the dataset $t$ into mini-batch: $D_t$.\\
    \textbf{2.} Define task-specific feature extractor model $M_t$ \\
    \textbf{3.} Initialize model parameters $\Theta_t$ randomly.  \\
}
Define shared feature extractor model $S$ and initialize their parameters $\Theta_s$ randomly.\\
Define the discriminator model $D$ and initialize their parameters $\Theta_d$ randomly.\\
Pre-train the discriminator. \\
 \For{$epoch$ in $1,2,...,epoch_{max}$}{
 \For{$t$ in $\{ t_1, t_2, \ldots, t_K \}$ }{
     \For{$batch$ in $D_t$}{
     \textbf{1.} TF $\leftarrow$ Generate the task-specific feature for \textit{batch} using model $M_t$\\
    \textbf{2.} SF $\leftarrow$ Generate the shared feature for \textit{batch} using model $S$\\
    \textbf{3.}  $\bar{h}= TF \oplus SF$\\
    \textbf{4.} Compute task-specific loss : $\mathcal{L}_{CE}^{t}$ \\
  \quad   \textit{// using Eq.(6)}\\
\textbf{5.} Compute adversarial loss : $\mathcal{L}_{adv}$  \\
\quad \textit{// using Eq.(13)}\\
    \textbf{6.} Update the parameters:
    \begin{equation}
    \nonumber
\begin{aligned}
     \Theta_t & = \Theta_t - \eta \frac{\partial \mathcal{L}_{CE}^{t}}{\partial \Theta_t} \\
     \quad \Theta_d & = \Theta_d - \eta \frac{\partial \mathcal{L}_{adv}}{\partial \Theta_d} \\
     \quad \Theta_s & = \Theta_s - \eta \big(  \frac{\partial \mathcal{L}_{CE}}{\partial \Theta_s}   -  \frac{\partial \mathcal{L}_{adv}}{\partial \Theta_s}\big) 
\end{aligned}
    \end{equation}
     }
 }
 }
 \caption{\label{algo:mt} Training Process}
\end{algorithm} 

\subsection{Results and Analysis}
Inspired by the recent success of the deep learning based frameworks in solving the relation extraction task, we develop three strong baselines based on STL and MTL frameworks for the purpose of comparison. Figure-2 provides the architecture for the below described baseline model. 
\begin{itemize}
\item \textbf{Baseline 1:} The first baseline is a STL model constructed by training a Bi-GRU on the features obtained from the embedding layer to capture the long-term dependencies as defined in Subsection-\ref{material_methods}. 
In our experiment, we build the individual model for each dataset.\\

\item \textbf{Baseline 2:} This single-task learning (STL) model is an advanced version of Baseline 1, where the sentence encoder of this model is also equipped with the word-level attention\cite{yang2016hierarchical}.
\item \textbf{Baseline 3:} 
It is a multi-task model with the shared Bi-GRU followed by word-level attention that acts as a shared feature extractor for all the tasks. 

\end{itemize}
\begin{table*}[h]
\centering
\begin{tabular}{l|l|l|l|l}
\hline
\textbf{System} & \textbf{Technique} & \textbf{P} & \textbf{R} & \textbf{F} \\ \hline
\hline
\textbf{Our System} & \textbf{Multi-task adversarial learning + GRU+attention} & \textbf{76.52}	& \textbf{69.01}	& \textbf{72.57}  \\ \hline
Joint AB-LSTM$^{*}$ \cite{sahu2017drug} & Bi-directional LSTM +attention & 71.84 & 66.88 & 68.99 \\ \hline
Yi et al.$^{+}$\cite{yi2017drug} & RNN+dynamic WE+ multi-attention & 73.67 & 70.79 & 72.20 \\ \hline
Joint AB-LSTM$^{+}$ \cite{sahu2017drug} & Bi-directional LSTM +attention & 73.41 & 69.66 & 71.48 \\ \hline
DCNN$^{+}$ \cite{liu2016dependency} & Dependency based CNN & 77.21 & 64.35 & 70.19 \\ \hline
Liu et al.$^{+}$ \cite{liu2016drug} & CNN & 75.72 & 64.66 & 69.75 \\ \hline
SCNN$^{+}$ \cite{zhao2016drug} & Two stage syntax CNN & 72.5 & 65.1 & 68.6 \\ \hline
\end{tabular}
\caption{Comparison with the SOTA techniques for DDI task. Performance is reported in terms of `P': Precision, `R': Recall, and `F': F1-Score (macro). $^{*}$ denotes the re-implementations of the systems proposed in [32] with the authors' reported experimental setups using their publicly available source code. $^{+}$ denotes that the reported results may not be directly comparable with our proposed system because of the difference in dataset statistics.}
\label{Comp-DDI}
\end{table*}
The results obtained by our proposed model and the baseline systems for each tasks is reported in Table-\ref{tab:eval-proposed}. The results obtained demonstrate the efficiency of our proposed MTL framework over other models that explore state-of-the-art techniques based on a single task and multi-task neural network.
For AiMed dataset, our proposed MTL model outperformed the Baseline $1$ and Baseline $2$ model by $6.96$ and $8.64$ F1-Score points, respectively. Similar trend was also followed for the BioInfer dataset where, we observe the performance improvements of $5.21$ and $7.58$ F1-Score points over Baseline $1$ and Baseline $2$, respectively, by the proposed approach. For DDI dataset, our proposed method attains improvements of $5.11$ and $4.97$ F1-Score points over Baseline $1$ and Baseline $2$, respectively. Lastly, for the i2b2-2010 clinical relation extraction dataset, the proposed method demonstrates the performance improvements of $1.56$ and $1.06$ F1-Score points over Baseline $1$ and Baseline $2$, respectively.\\
We observe that our model outperforms Baseline 3 model, with $3.06$, $1.72$, $2.86$, and $0.94$ F-Score points on AiMED, BioInfer, DDI, and i2b2-2010 clinical relation extraction dataset respectively.\\
Overall, the findings achieved show that using self attention based adversarial multi-task learning to save the present knowledge in a shared layer is helpful for a new task.\\
\textbf{Ablation Study:}
To analyze the impact of various component of our model, we perform the ablation study by removing one component from the proposed model and evaluate the performance on all the four tasks. We carried of two set of ablation study: (1) jointly training all the four dataset together (c.f. Table-\ref{ablation}), and (2) training two similar task dataset i.e. AiMed and BioInfer together (since both share characteristics of having protein-interaction information ), DDI and i2b2-RE task together (c.f. Table-\ref{ablation-1}). One possible reason for the i2b2-RE dataset to not be benefited from adversarial learning component is firstly because it is jointly trained with DDI corpus whose origins (clinical notes vs biomedical literature) and characteristics are not very similar to the i2b2-RE dataset in contrary to AiMED and Bionfer that are both protein-interacting datasets. 
Another reason is the huge difference between the samples in DDI ($5550$) and i2b2-RE ($62254$). In this case when the batches of the DDI dataset finish arriving, model will train only with the i2b2-RE dataset and could lead to the diminishing effect of the adversarial training.\\
Experimental results shows that removing self attention leads to the decrements in the performance of the model across all the task, in both the ablation study scenarios. However, we observe that adversarial learning was not that much effective in the ablation study 1, where we jointly trained all the four dataset together. While for the ablation study 2, when we trained similar task together, there was a drop in the performance across all the dataset. This shows that, adversarial learning is much helpful in the related task where the data distribution is similar.
\subsection{Comparative Analysis}
In this section, we will conduct the comparative analysis of our proposed method with the state-of-the-art (SOTA) model for all the three tasks.
\begin{itemize}
\item \textbf{Protein-Protein Interaction:} We used the SOTA methods for both datasets as presented in Table-\ref{results} to compare our proposed model.
The results show that the proposed model significantly outperforms the SOTA systems. From this we can conclude that our proposed multi-task model is more potent in extracting interacted protein pairs over the architecture based on CNN established in \cite{hua2016shortest} and LSTM framework \cite{hsieh2017identifying}. 
Our adversarial MTL model achieves a significant $4.26$ F-score point increment over the LSTM based model \cite{hsieh2017identifying} on AiMed dataset. 
In case of BioInfer dataset, our proposed model was able to achieve significant performance improvement of $3.09$ F-Score point over \cite{hsieh2017identifying} as shown in Table-\ref{results}. However, in comparison to \cite{hua2016shortest}, we could observe the improvement of $1.13$ F-Score points over the proposed model. This clearly demonstrates the effect of neural self-attention based adversarial learning in multi-task setting.
\item \textbf{Drug-Drug Interaction}
We compare our proposed model with SOTA DDI extraction techniques as shown in Table-\ref{Comp-DDI}. Since, there was a difference in the dataset statistics, we re-implemented the system proposed by \cite{sahu2017drug} using their publicly available source code on our DDI extraction dataset. Our multi-task adversarial Att-LSTM model obtains the significant performance improvement of $3.58$ F-Score points over the state-of-the-art system \cite{sahu2017drug} that exploited Bi-LSTM with attention mechanism. 
Also, the obtained experimental results illustrate that the model for the DDI task is benefited from other similar tasks, more specifically from the PPI tasks. It is because of the high semantic similarity between the sentences in DDI and PPI tasks.
\item \textbf{Medical Concept Relation}
We were unable to make direct comparison of our proposed approach with the systems participated in i2b2-2010 shared task due to the incomplete dataset. We compare our model with \cite{sahu2016relation}, as they also experimented with the same dataset. The results are reeported in Table \ref{i2b2-comp},. We obtain the performance improvements of $12.51$, $9.91$, $5.04$, $11.39$, and $5.65$ F-Score points over \cite{sahu2016relation} (irrespective of the use of additional linguistic features) for TeCP, TrCP, PIP, TrAP, and TeRP classes, respectively.
This shows the usefulness of self attention based adversarial learning in multi-task setup which eventually gathers the complimentary features for medical concepts relations and other related tasks and improves the performance of the system.
\end{itemize}
\begin{table}[!h]
\centering
\begin{tabular}{l|l|l|l|l}
\hline
\textbf{System} & \textbf{Sahu et al. \cite{sahu2016relation}} & \multicolumn{3}{l}{\textbf{Proposed Approach}} \\ \hline
\textbf{Relation-types} & \textbf{F} & \textbf{P} & \textbf{R} & \textbf{F} \\ \hline
\textbf{TeCP} & 50.56 & 62.89  & 63.43 & \textbf{63.07} \\ \hline
\textbf{TrCP} & 56.44 & 68.44 &64.52  & \textbf{66.35} \\ \hline
\textbf{PIP} & 64.92 & 73.20 & 67.17 & \textbf{69.96} \\ \hline
\textbf{TrAP} & 69.23 & 79.10 & 82.28 & \textbf{80.62} \\ \hline
\textbf{TeRP} & 81.25 & 87.31 & 86.56 & \textbf{86.90} \\ \hline
\end{tabular}
\caption{Comparison with the SOTA techniques for i2b2-2010 clinical relation extraction task. For the fair comparison, we have reported the weighted F-Score.}
\label{i2b2-comp}
\end{table}


\begin{figure*}[t!]
  \centering
  \begin{subfigure}[t]{0.32\textwidth}
    \includegraphics[height=1.5in, width=\textwidth]{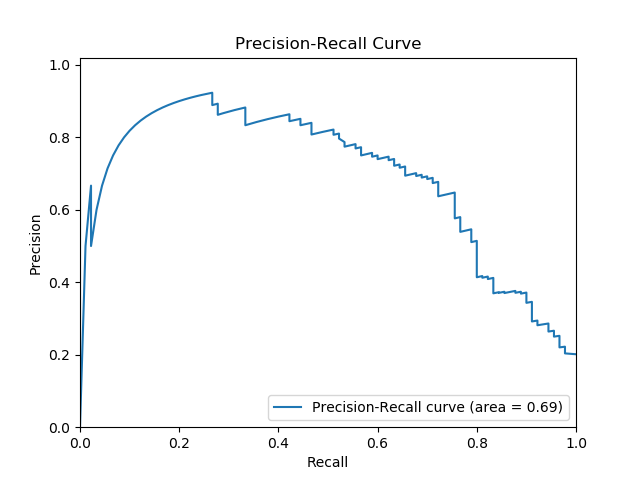}
     \caption{PR curve for AiMed}
  \end{subfigure}
  \begin{subfigure}[t]{0.32\textwidth}
    \includegraphics[height=1.5in, width=\textwidth]{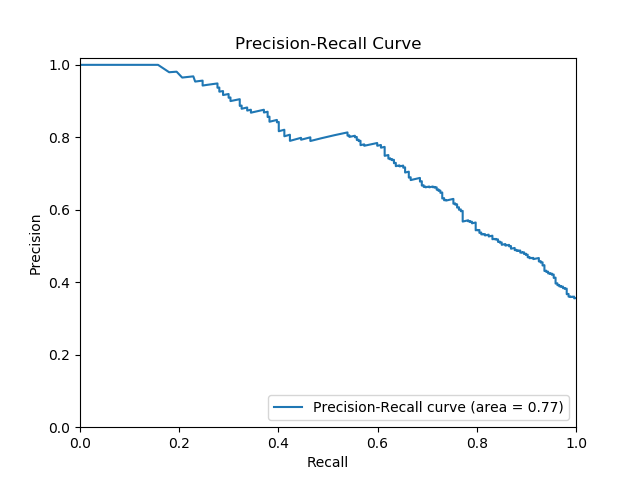}
    \caption{PR curve for BioInfer}
  \end{subfigure}
 \caption{The graphs \textbf{(a)} and \textbf{(b)} demonstrate the Precision-Recall curves for the binary classification dataset, AiMed and BioInfer.}
  \label{learning-curve}
\end{figure*}
\begin{figure*}[h]
\centering
\includegraphics[width=\linewidth, height=3cm]{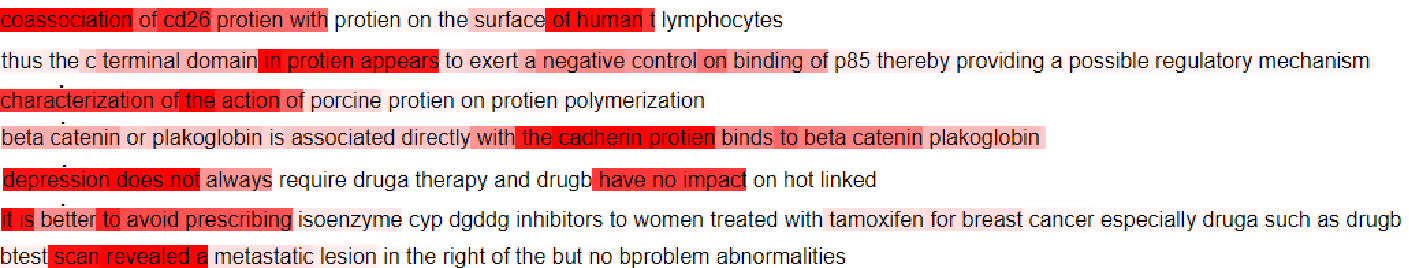}
\caption{Heatmap of the attention weight distributions on the examples of different datasets. The intensity of the colour is increased with the increment in the attention weight.}
\label{atten-visual}
\end{figure*}

\subsection{Error Analysis}
Here, we closely examine the various forms of errors with respect to the tasks that cause the mis-classification.

\begin{enumerate}
\item \textbf{PPI:} In case of the AiMED and BioInfer datasets, we observe that in a sentence with multiple protein mentions, our proposed model fails to identify properly the interacted pair. For example:\\

\textbf{Sentence 1: }``\textit{We screened proteins for interaction with PROTEIN and cloned the full-length cDNA of human PROTEIN which encoded 1225 amino acids.}"\\

Here, the actual label was true but our model predicted this as false. Repetitive mentions of proteins behave like a noise, than can inhibit the model to extract contextually relevant information.
We also made an interesting observation that in the presence of the protein interacting words (such as `\textit{bind}', `\textit{interact}'), our model predicts the class label as `\textit{interacting}' (true). For example:\\

\textbf{Sentence 1: }``\textit{PROTEIN binding significantly increased hetero- and homo-oligomerization (except for the BR-II homo-oligomer, which binds ligand poorly in the absence of PROTEIN}"\\

\textbf{Sentence 2: }``\textit{PROTEIN is a muscle-specific HLH protein that binds DNA in vitro as a heterodimer with several widely expressed HLH proteins, such as the PROTEIN gene products E12 and E47.}"\\

This is because these words often occur in the vicinity of the interacting protein mentions. \\
\item \textbf{DDI:} Apart from the highly imbalance dataset issue, our model fails to capture the exact relationship between the DDI pairs, where the lengths of the sentences were long. Another phenomenon that we observed as a source of mis-classification was that the ``\textit{Int}'' type was often predicted as ``\textit{Effect}'' class type. ``\textit{Int}'' class describes the coarse classification, i.e., there exists interaction between two drugs. This implies that there could be a positive or negative outcome which forms the main cause of the system often getting 
confused between the ``\textit{Int}'' and ``\textit{Effect}'' class labels. For example:\\

\textbf{Sentence 1: }``\textit{Synergistic interaction between DRUGA and DRUGB is sequence dependent in human non small lung cancer with EGFR TKIs resistant mutation.}" \\
We also found that some labels were incorrectly predicted because of class-specific keywords which exist in the sentence but are not related to the concerned entity pair. For example:\\

\textbf{Sentence 1: }``\textit{Interaction study of DRUGA and DRUGB with co administered drugs .}"\\

\textbf{Sentence 2: }``\textit{If in certain cases , an DRUGA is considered necessary , it may be advisable to replace tamoxifen with DRUGB.}"\\

In sentence 1, the model got confused between the class label `Int' and `None', while in the sentence 2, our model incorrectly predicted the class label as `Advice'.\\
\item \textbf{Medical Concept Relation:} We observe that due to close similarity between the class label `\textit{TeRP}' and `\textit{TeCP}', our model was found to be confused between these classes. For example:\\

\textbf{Sentence 1: }``\textit{TEST x-ray revealed no PROBLEM , no congestive heart failure.}"\\

In the given sentence, our model incorrectly predicted it as class `\textit{TeRP}'.\\
We also observed that majority of the misclassification was between `\textit{PIP}' and `\textit{NONE}' class. For example:\\

\textbf{Sentence 1: }``\textit{There was PROBLEM atrophy and PROBLEM encephalomalacia.}" 

\end{enumerate}
\subsection{Visual Analysis:} We have carried out the visual analysis (in Figure \ref{atten-visual}) to get an intuitive understanding of attention weights in multi-task attention model. Each sentence in the figure shows the attention distribution in the form of heatmap for an instance of the corresponding dataset. The highlighted colours indicate the most relevant words in the sentence selected by the attention mechanism. For example in the sentence, ``\textit{co-association of cd26 protein with the protein on the surface of human t lymphocytes}", the model is able to provide more weights to ``\textit{co-association of cd26 protein}", which is relevant to correctly classify the given protein pair as interacted pair. 
 For the binary classification datasets, AiMed and BioInfer, we also plot the Precision-Recall curve (c.f. Figure-\ref{learning-curve}). 
\section{Conclusion}
In this paper, we propose a unified multi-task learning approach that exploits the capabilities of adversarial learning approach for relation extraction from biomedical domain. We first experimented on three benchmark biomedical relation extraction tasks, i.e., protein-protein interaction, drug-drug interaction, and clinical relation extraction. For that, we utilized four popular datasets: AIMed, BioInfer, SemEval 2013 DDI shared task dataset and i2b2-2010 clinical relation dataset. We demonstrated that our model shows superior performance compared to state-of-the-art models for all the tasks. \\
Although our model has shown significant improvements over state-of-the-art methods on all the tasks, it was observed that our supervised model does not generalize well for the class label with the small instances. In future, we would like to develop a zero-shot learning method that could assist the model in the huge class imbalance issue. 

{\bf Acknowledgement}: 
Dr. Sriparna Saha gratefully acknowledges the Young
Faculty Research Fellowship (YFRF) Award, supported by
Visvesvaraya Ph.D. Scheme for Electronics and IT, Ministry of
Electronics and Information Technology (MeitY), Government
of India, being implemented by Digital India Corporation
(formerly Media Lab Asia) for carrying out this research.

\bibliographystyle{IEEEtran}
\bibliography{document}
\end{document}